\documentclass[conference]{IEEEtran}
\IEEEoverridecommandlockouts
\usepackage[utf8]{inputenc}
\usepackage[english]{babel}
\usepackage{cite}

\usepackage{hhline}
\usepackage{amsmath,amssymb,amsfonts}
\usepackage[H]{algorithm}
\usepackage{algorithmic}
\usepackage{graphicx}
\ifCLASSOPTIONcompsoc
	\usepackage[caption=false, font=normalsize, labelfont=sf, textfont=sf]{subfig}
\else
	\usepackage[caption=false, font=footnotesize]{subfig}
\fi
\usepackage{booktabs}
\usepackage{textcomp}
\usepackage{xcolor}

\title{Training DNN IoT Applications for Deployment On Analog NVM Crossbars}

\author{
\IEEEauthorblockN{García-Redondo, Fernando}
\IEEEauthorblockA{\textit{Arm Ltd. Cambridge} \\
fernando.garciaredondo@arm.com}
\and
\IEEEauthorblockN{Shidhartha Das}
\IEEEauthorblockA{\textit{Arm Ltd. Cambridge} \\
shidhartha.das@arm.com}
\and
\IEEEauthorblockN{Glen Rosendale}
\IEEEauthorblockA{\textit{Arm Ltd. San José} \\
 glen.rosendale@arm.com}
}

\begin{document}

\bstctlcite{IEEEexample:BSTcontrol}

\maketitle

\begin{abstract}
A trend towards energy-efficiency, security and privacy has led to a recent focus on deploying deep-neural networks (DNN) on microcontrollers.
However, limits on compute and memory resources restrict the size and the complexity of the machine-learning (ML) models deployable in these systems. Computation-In-Memory architectures based on resistive nonvolatile memory (NVM) technologies hold great promise of satisfying the compute and memory demands of high-performance and low-power, inherent in modern DNNs.
Nevertheless, these technologies are still immature and suffer from both the intrinsic analog-domain noise problems and the inability of representing negative weights in the NVM structures, incurring in larger crossbar sizes with concomitant impact on Analog-to-Digital Converters (ADCs) and Digital-to-Analog Converters (DACs).
In this paper, we provide a training framework for addressing these challenges and quantitatively evaluate the circuit-level efficiency gains thus accrued. We make two contributions:  Firstly, we propose a training algorithm that eliminates the need for tuning individual layers of a DNN ensuring uniformity across layer-weights and activations. This ensures analog-blocks that can be reused and peripheral hardware substantially reduced. Secondly, using Network Architecture Search (NAS) methods, we propose the use of unipolar-weighted (either all-positive or all-negative weights) matrices/sub-matrices. Weight unipolarity obviates the need for doubling crossbar area leading to simplified analog periphery. We validate our methodology with CIFAR10 and HAR applications by mapping to crossbars using $4$-bit and $2$-bit devices. We achieve up to $92.91\%$ accuracy ($95\%$ floating-point) using $2$-bit only-positive weights for HAR.
A combination of the proposed techniques leads to $80 \%$ area improvement and up to $45 \%$ energy reduction.
\end{abstract}


\section{Introduction}

The deployment of DNN applications on always-ON
IoT devices suffers from stringent limitations on memory,
computation capabilities and memory-bandwidth \cite{Zidan2018}, leading to complex trade-offs between model-size, performance and energy-consumption.
This trade-off is addressed in the prior-art through a combination of algorithmic approaches (network-pruning and weight elision, quantization \cite{Kodali2017, Fedorov}), system-design solutions (NN accelerators, vectorized instructions \cite{Huang2017a, Zhang2017b, Kodali2017, Whatmough2019}) and process-technology innovations (3D-integration, emerging non-volatile memory technologies 
\cite{Huang2017a, Cai2019}). 

\emph{Computation In Memory (CIM)} architectures
drastically reduce memory-bandwidth requirements \cite{Zidan2018, Hamdioui2019},
while taking advantage of the quantization and pruning solutions.
Emerging resistive switching technologies such as \emph{Phase-Change Memory (PCM)}
and memristors or \emph{Resistive Random Access Memories (RRAM)}
behave as analog synapses placed in crossbar arrays
\cite{Li2017e, Rahimi2017,Ambrogio2018, Serb2018, Hamdioui2019, Joshi2019}.
%
%
By executing the multiply-accumulate (MAC) operation in these highly-integrated structures in the analog-domain, NVM crossbars naturally provide the parallelization of the matrix-vector multiplication, the kernel behind most DNN operations.
%
Moreover, as the weights are encoded in the non-volatile resistive elements and not transferred from offchip external memory, energy consumed due to data movement is significantly reduced, potentially enabling orders of magnitude
improvements in energy and computing efficiency.

Despite the promise of performance and energy consumption improvements, the stark reality remains that emerging NVM technologies are still relatively immature and suffer from intrinsic analog-domain shortcomings such as device variability, sensitivity to temperature variations and no intrinsic capability for representing negative values.
%
These constraints are considerable technological challenges that need to be overcome before CIM architectures using NVM can be deployed in mass market. 

Addressing these substantial challenges entirely through innovative circuit-design can severely dent potential efficiency gains. As an illustrative example, mapping negative weights in NVM devices requires independent analog-domain accumulators for positive and negative weights, thereby potentially doubling the analog-domain circuit overheads, including energy-expensive ADCs and DACs \cite{Li2017e, Joshi2019}. Similarly, differences in the dynamic range of the activations across the DNN layers require per-layer tuning of analog periphery.
This limits the opportunities for reusable analog macros
\cite{Serb2018}, and creates a significant gap between laboratory prototypes that rely upon external probes and analyzers
\cite{Li2017e, Ambrogio2018, Li2018a}, and deployable market products. 

In this work we introduce a framework to train and efficiently map the DNN to the NVM hardware, providing two main contributions. Firstly,
we enable the re-use of smaller, less power-hungry, uniform
analog modules across different layers in the DNN, removing 
the need of per-layer full-custom periphery design.
By ensuring uniform scaling through layers, independently of their morphology and size,
area and power benefits come together with
shorter circuit design time, closing the gap between
reconfigurable blocks \cite{Li2017e, Serb2018} and real reconfigurable solutions.
Secondly, we investigate how relaxing bipolar-weight matrices requirements
can lead to additional periphery area savings, while reducing the crossbar area by $2 \times$.
We prove that certain applications can be retrained and directly mapped to unipolar weight matrices.
Conversely, for those deeper convolutional NN that cannot be completely unipolar,
we analyze the trade-off between the number of unipolar channels (and therefore the energy/area savings) and the accuracy loss.

The paper is structured as follows.
Section \ref{sec:motivation} deepens into the motivation and related work.
Section \ref{sec:hw_aware} describes the proposed methodology and framework,
followed by the achieved results on Section \ref{sec:results}.
Finally, we summarize and conclude this paper in the conclusion section.

\section{Related Work}
\label{sec:motivation}

\subsection{NVM Crossbars as MAC Engines}
\begin{figure}[!t]
\centering
\includegraphics[width=\columnwidth]{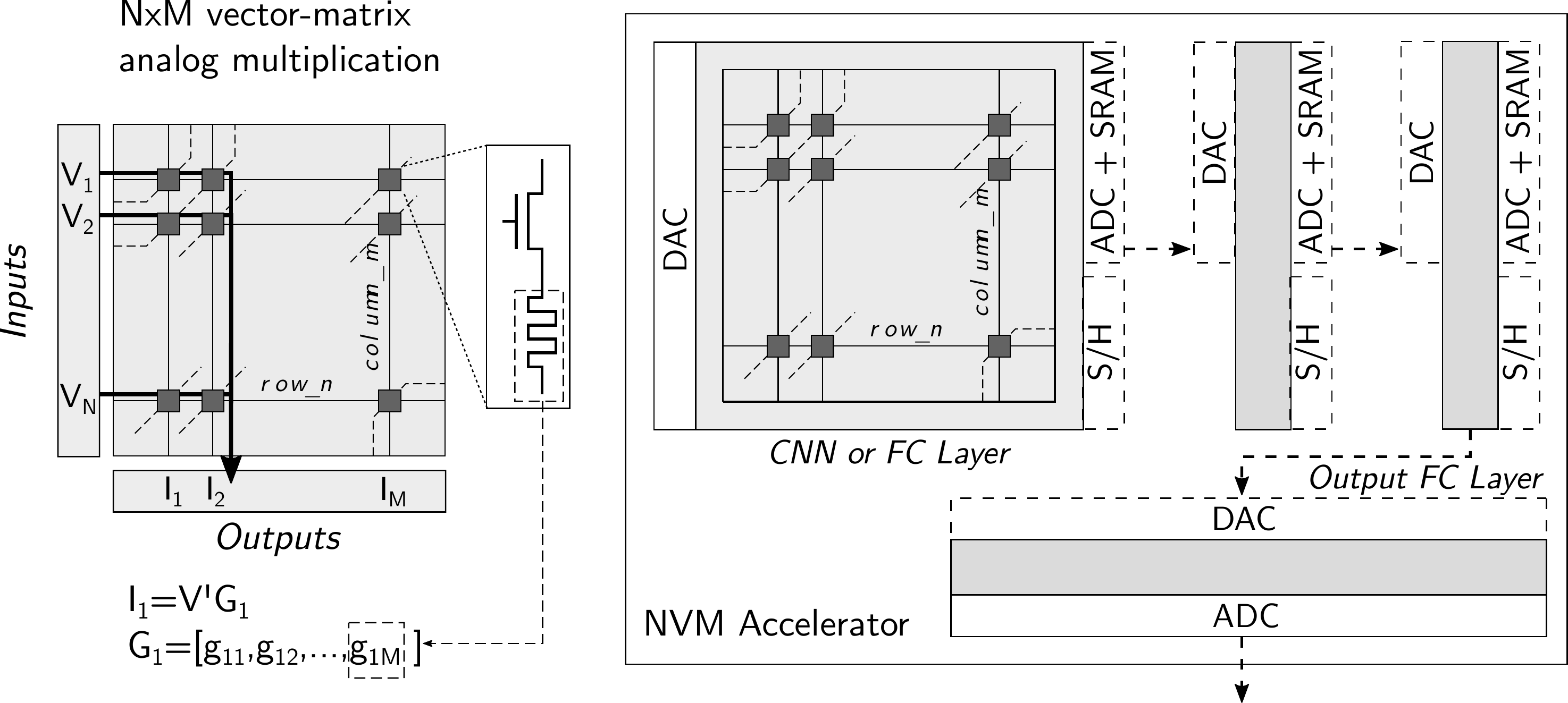}
\caption{
	NVM accelerator architecture
	and principle behind the NVM based vector-matrix multiplication.
	Inputs are encoded as voltages, and drive the crossbar through the
	horizontal row lines.
	Bitline currents naturally accumulate the individual products of
	the input voltages and the NVM conductances and are later digitized.
	In very deep neural-networks (NN) crossbars are interconnected digitally.
}
\label{fig:nvm_accelerator}
\end{figure}

\begin{figure*}[!t]
\centering
\includegraphics[width=\textwidth]{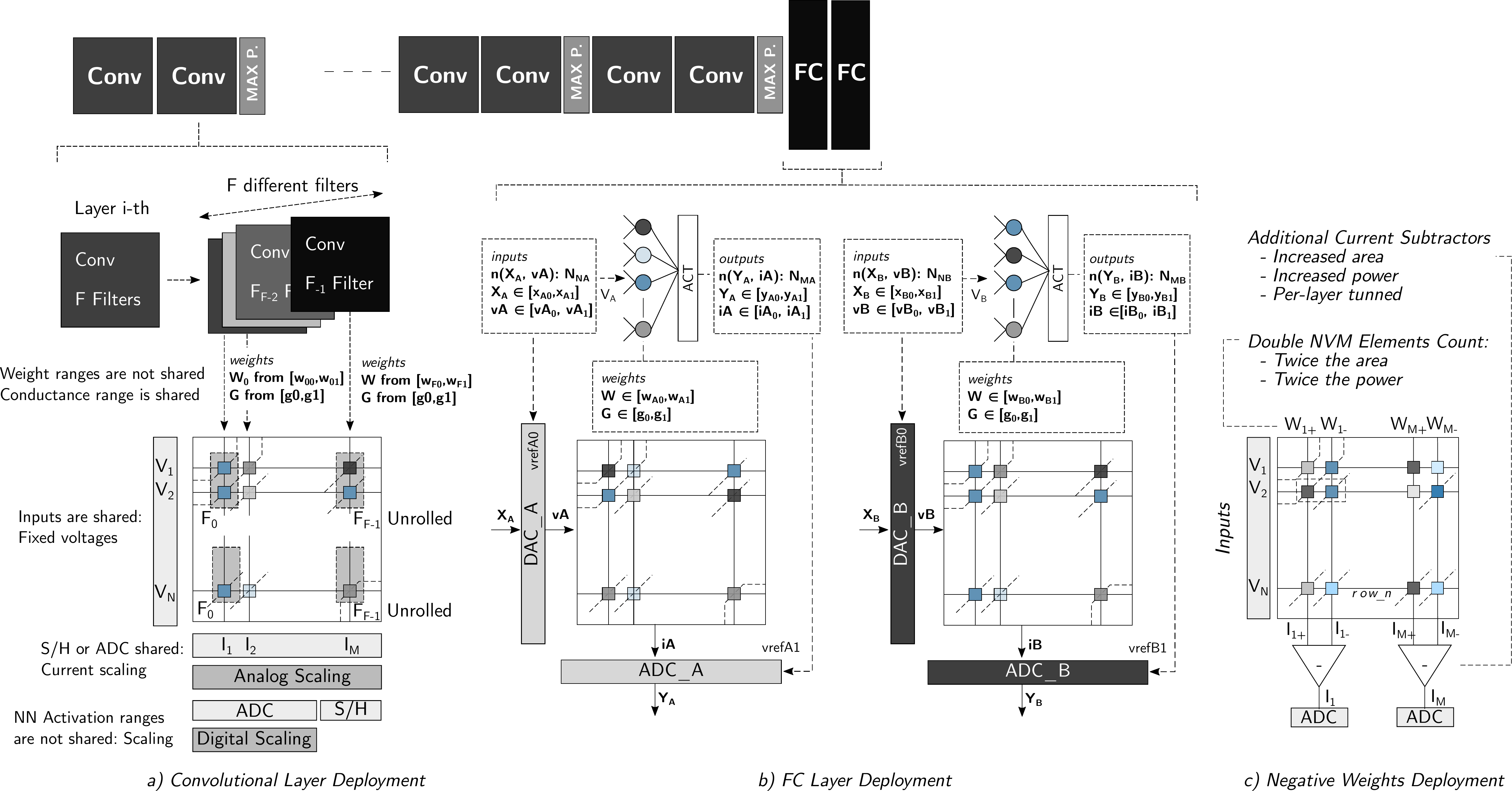}
\caption{
	Common problems on the deployment of NN layers in NVM crossbars:
	Scaling isues on \emph{a)} convolutional layers and
	\emph{b)} fully connected layers are often solved using full-custom periphery,
	while the handling of \emph{c)} negative weights area and power overcosts.
}
\label{fig:nvm_problems}
\end{figure*}

Machine Learning applications, and more specifically DNN,
use vector-matrix multiplication operations (also called multiplication-accumulation or \emph{MACs} operations) as a common underlying primitive for most algorithmic operations. Resistive NVM elements (PCM, RRAM, MRAM) arranged in crossbar topologies compute MAC operations in constant time with significantly improved energy-efficiency
\cite{Rahimi2017, Ambrogio2018, Hamdioui2019, Joshi2019} by mapping the fundamental MAC operation to the analog-domain. Thus, an individual element in the weight matrix is mapped to an individual NVM element whose conductance value is programmed to a discrete conductance $g$ within a known range $g \in [g_{ON}, g_{OFF}]$. By encoding second operand as a voltage $v$, the current through the device becomes the multiplication of both operands $i = v g$.

The crossbar architecture automatically computes the addition of the individual dot-products.
As depicted in Figure~\ref{fig:nvm_accelerator}, the set of accurately programmed conductances
in the NVM devices conforms to the matrix $G=\{g_{ab}\}, a=[1,N], b=[1,M]$.
Encoding the input vector as voltages $V = \{v_a\}$,
the current flowing through each one of the bitlines $I = \{i_b\}$
corresponds to the accumulation of the
partial products $i_b = \sum_{a=1}^{N}{v_a g_{ab}}$.

\subsection{NVM Technology Level Challenges}
García-Redondo et al~\cite{Garcia-Redondo2017} provide a more detailed perspective on resistive switching NVMs. Three key challenges are: endurance, variability, and device non-linearity.
Limited endurance after many writes is one of the main problems that remains unsolved.
As an example, in-crossbar training methods have been proposed~\cite{Ambrogio2018},
but their applicability to real products is still unclear due to this reduced lifetime.
Always-ON inference applications heavily rely on analog read operations, 
and rarely re-write the weights encoded in the crossbar. Consequently limited write-ability
would not affect the normal behavior.
Second, variability and crossbar-related errors heavily affect NVM-CMOS hybrid circuits.
Nevertheless architectures as DNNs are naturally robust against the noise that these problems may cause.
More over, this defects can be taken into account at training time,
getting around device faults \cite{Hasan2017}.
Third, NVM elements suffer from non-linearities that lead to errors during the weight-to-conductance mapping.
However, by engineering the physical device this problem can be overcome. 

As seen, technology problems related to inference ML accelerators built with
NVM technologies can be overcome.
Next section describes the challenges specific to the deployment of ML algorithms in NVM crossbars.

\subsection{NVM For Analog ML Accelerators: Circuit Challenges}
In this section the three main challenges still to be addressed at circuit level: 
operands precision, analog signals dynamic range control and negative weights representation.

\subsubsection*{Challenges Related to Precision}
For both digital and emerging analog accelerators,
the precision of the operands and operations involved in the DNN determines
the accuracy, latency and energy consumption of the inference operation.
Consequently, the quantization of both weights and activations is critical
on the design of the accelerated system.
Though $6-8$-bit NVM devices have been demonstrated \cite{Li2017e},
variability or analog noise may compromise the encoding of
more than $2$ to $4$ bits per cell/weight.
On the other hand, the current accumulation taking place on the column bit-line
is not quantized and does not suffer from precision related problems.
Finally, the precision of the involved DACs and ADCs 
greatly influences the total area and power consumption \cite{Li2017e, Ni2017}.
Thereby, the selection of the periphery precision (or the design/use of multiple DACs/ADCs exhibiting
different bits) is critical for the system accuracy and efficiency.

\subsubsection*{Challenges Related to Dynamic Range}
The dynamic range of the analog signals decides the periphery design and reconfigurability.
Figure \ref{fig:nvm_problems} describes how NN layers are deployed in an NVM crossbar.
Convolution kernels are decomposed by sets of channels, and mapped to different columns.
Then, the kernels are unrolled and grouped to compute in parallel the convolution operation.
However, the input activations, weight, accumulation signals and output activations
do not share common ranges across different channels and layers.
This \emph{non-uniform scaling} across DNN layers imposes full-custom blocks per stage
--Figure~\ref{fig:nvm_problems} \emph{a)} and \emph{b)}.
The number of elements in a layer and the range of the involved signals
determine the currents flowing through the bitlines.
Different voltage or current signals require per-layer full-custom periphery.

To the best of our knowledge, every \emph{offline learning} work
in the literature that trains the DNN externally to the NVM crossbar
dynamically scale each DNN layer to the available set of
conductances in which we can program the device \cite{Hu2016c}.
This process is independent of the input, weight and activation value ranges of the layers.
However, a reconfigurable system deals with variable voltage/current signals with very different ranges:
though the crossbar weights
can be reprogrammed, the full-custom periphery constraints the deployment of different NN graphs to only one.
Consider the deployment of two different fully connected layers, $A$ and $B$, of the NN described in Figure~\ref{fig:nvm_problems} \emph{b)}.
The number of inputs $n(X_A)$ and neurons $n(Y_A)$ differs from those in layer $B$.
Similarly, their ranges $[x_{A0}, x_{A1}]$, $[y_{A0}, y_{A1}]$ will differ from
the respective ones in layer $B$.
And more importantly, the weight matrices $W_A$ and $W_B$ differ on their
ranges $[w_{A0}, w_{A1}]$, $[w_{B0}, x_{B1}]$.
However, both matrices $W_A$ and $W_B$ need to be mapped \cite{Hu2016c} to
the same available set of conductances the devices can be programmed in, $G$.
Thus for translating the $i-th$ weight matrix to the
conductances range $[g_0, g_1]$, the periphery generating the required voltage
amplitudes $v_{ith}$ and sensing the output currents $i_{ith}$
needs to be scaled accordingly, and therefore be different.
Full-custom blocks require higher design time \cite{Giordano2019}, and limit the deployment of
different NNs in the same HW. Moreover, should the NN weights be updated varying
voltage/current ranges, DACs/ADCs would require extra calibration processes.

Similarly, and to take fully advantage of the crossbar, the deployment of convolutional layers
requires mapping different filters of the same layer to different columns in the tile.
As described in Figure~\ref{fig:nvm_problems} \emph{a)},
\emph{per-channel} quantization methods \cite{Krishnamoorthi2018} lead to different
weight/activation ranges. As voltage inputs and S/H or ADC elements are shared
across the filters, analog/digital scaling stages would be required.
Different S/H and ADC designs leads to additional area, power consumption, and higher design times \cite{Giordano2019}.

\subsubsection*{Challenges Related To Weights Polarity}

The conductance in a passive NVM element can only be a positive number $g$ in the range $[g_{OFF}, g_{ON}]$.
However, the NN weights, no matter whether $W \in \mathbb{R}$ or $W \in \mathbb{Z}$, contain both positive and negative values.
Consequently, the use of bipolar weights involves a problem when mapped to
a only positive conductance set.
Traditionally positive and negative weights
are deployed separately in different areas of the crossbar \cite{Li2017e, Joshi2019}.
This approach comes with the duplication of crossbar area and energy consumption,
and the addition of current subtractors or highly-tuned differential ADC --hindering the reconfigurability of the accelerator.
As depicted in Figure~\ref{fig:nvm_problems} c), using this scheme, we double the crossbar area as per-weight, one column computes
the positive contributions, while the other column the negative ones \cite{Li2017e, Joshi2019, Cai2019}.
Moreover, additional current subtraction blocks are required before/at the ADC stages \cite{Joshi2019, Cai2019}.
Alternative solutions as \cite{Hu2016c} shifting the weight matrices usually involve
the use of biases dependent on the inputs and additional periphery.
Nevertheless, both alternatives involve considerable area\&energy overheads.

\subsubsection*{Status of NVM-based Reconfigurable Accelerators}

A naive deployment of a particular algorithm into a given crossbar
requires the periphery surrounding it to be full-custom designed.
Therefore, despite many efforts have been devoted to design NVM based accelerators,
most works presented in literature describing different NN experiments
rely on HW external to the chip to assist the crossbar as supporting periphery
\cite{Hu2016c, Ambrogio2018, Li2018a, Joshi2019}.
To solve the issue, the first reconfigurable CMOS-NVM processor includes
per-column current dividers as scaling stages,
interfacing the high-precision ADCs before the conversion takes place \cite{Cai2019}.
While achieving reconfigurability,
the system is penalized in terms of area and power.

\section{Hard-Constrained HW Quantized Training}
\label{sec:hw_aware}

To address the reconfigurability versus full-custom periphery design, and
its dependence on the weights/activation precision,
we have developed a framework to aid mapping the DNN to the NVM hardware at training time.
The main idea behind it is the use of hard-constraints when computing
forward and back-propagation passes.
These constraints, related to the HW capabilities, impose the precision used
on the quantization of each layer, and guarantee that
the weight, bias and activation values that each layer can have
are shared across the NN. 
This methodology allows, after the training is finished, to
map each hidden layer $L_i$ to uniform HW blocks sharing:
\begin{itemize}
\item a single DAC/ADC design performing $\mathcal{V}()$ / $act()$ 
\item a single weight-to-conductance mapping function $f()$
\item a global set of activation values $Y_g = [y_0, y_1]$
\item a global set of input values $X_g = [x_0, x_1]$
\item a global set of weight values $W_g = [w_0, w_1]$
\item a global set of bias values $B_g = [b_0, b_1]$.
\end{itemize}
Being the crossbar behavior defined by
\begin{align}
	i_{ij} &= \sum v_{ik} g_{ikj} + b_{ij} \\
 	v_{ik} &= \mathcal{V}(x_{ik}) \\ 
	g_{ikj} &= f(w_{ikj}) \\
	y_{ij} &= act(i_{ij}),
\end{align}
and every system variable within the sets
$Y_g, X_g, W_g$ and $B_g$,  
every DAC/ADC performing $\mathcal{V}()$ and $act()$
will share design and can potentially be reused.
To achieve the desired behavior we need to ensure at training time that the following
equations are met for each hidden layer $L_i$ present in the NN:
\begin{align}
 Y_i &= \{y_{ij}\}, y_{ij} \in [y_0, y_1] \\
 X_i &= \{x_{ik}\}, x_{ik} \in [x_0, x_1] \\
 W_i &= \{w_{ikj}\}, w_{ikj} \in [w_0, w_1] \\
 B_i &= \{b_{ij}\}, b_{ij} \in [b_0, b_1].
\end{align}

Commonly, the output layer activation
(\emph{sigmoid, softmax}) does not match the hidden layers activation.
Therefore for the DNN to learn the output layer should be quantized using
an independent set of values $Y_o, X_o, W_o, B_o$ that may or not match $Y_g, X_g, W_g, B_g$.
Consequently, the output layer is the only layer that once mapped to the crossbar
requires full-custom periphery.

\subsection{HW Aware Graph Definition}
\begin{figure}[!t]
\centering
\includegraphics[width=\columnwidth]{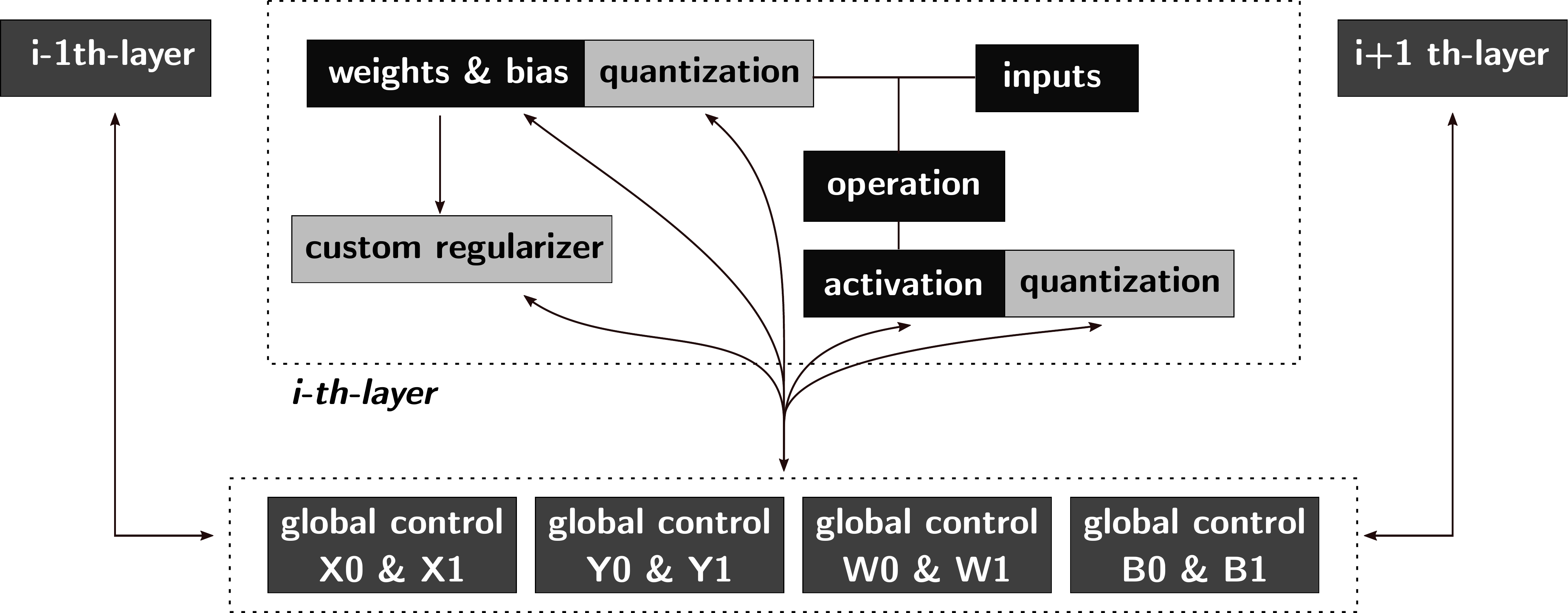}
\caption{
	Simplified version of the proposed quantized graph for crossbar-aware training,
	automatically handling the global variables
	involved in the quantization process, achieving uniform scaling across layers.
}
\label{fig:training_quant_graph}
\end{figure}
The NN graphs are generated by \emph{Tensorflow Keras} libraries.
In order to perform the HW-aware training, elements controlling the quantization,
accumulation clippings, and additional losses, are added to the graph.
Figure \ref{fig:training_quant_graph} describes these additional elements, denoted as
\emph{global variables}. For this purpose, 
the \emph{global variable control} blocks manage the definition,
updating and later propagation of the \emph{global variables}.
A \emph{global variable} is a variable used to compute a global set of values $V_g$
composed of the previously introduced $Y_g, X_g, W_g, B_g$ or others.
\emph{Custom regularizer} blocks may also be added to help
the training to converge when additional objectives are present.

\subsection{HW Aware NN Training}
\subsubsection{Differentiable Architecture and Variables Updating During Training}
Each \emph{global variable} can be non-updated during training,
--fixing the value of the corresponding
global set in $V_g$-- or dynamically controlled using the related \emph{global variable control}.
If fixed, a design space exploration is required in order to find the best set of
\emph{global variable} hyperparameters for the given problem.
On the contrary, we propose the use of a \emph{Differentiable Architecture (DA)} \cite{Liu2018b}
to automatically find the best set of \emph{global variable} values using the back-propagation.
In order to do that, we make use of DA to explore the NN design space.
To achieve it, we define the \emph{global variables} as a function of each layer
characteristics --mean, max, min, deviations, etc.
If complying with DA requirements, the global control elements automatically
update the related variables descending through the gradient computed
in the back-propagation stage.
On the contrary, should a specific variable not be directly computable
by the gradient descent,
it would be updated in a later step as depicted in algorithm~\ref{alg:darts}.
\begin{algorithm}[tb]
\caption{Quantized Training aided by Differentiable Architecture Search \cite{Liu2018b}}
\label{alg:darts}
\begin{algorithmic}
	\STATE {\bfseries Input:} Set of global variables $V_g = \{X_g, Y_g, W_g, B_g\}$
	\STATE Initialize $V_g$
	\WHILE{not converged}
	\STATE Update weights $W$
	\STATE Compute non-differentiable  vars in $V_g$
	\STATE Update layer quantization parameters
	\ENDWHILE
\end{algorithmic}
\end{algorithm}

We also propose the use of DA on the definition of inference networks that
target extremely low precision layers (i.e. $2$ bit weights and $2-4$ bits in activations),
to explore the design space, and to find the most suitable activation functions
to be shared across the network hidden layers.
In Section~\ref{sec:results} experiments we explore the use (globally, in every hidden layer) of
a traditional \emph{relu} versus a customized \emph{tanh} defined as $tanh(x - th_g)$.
Our NN training is able to choose the most appropriate activation, as well as
to find the optimal parameter $th_g$.
The parameter $th_g$ is automatically computed through gradient descent.
However, to determine which kind of activation to use, we first define the
continuous activations design space as
\begin{equation}
act(x) = a_0 relu(x) + a_1 tanh(x - th_g),
\end{equation}
where $\{a_i\} = \{a_0, a_1\} = A_g$.
The selected activation $a_s$ is obtained after applying \emph{softmax} function on $A_g$:
\begin{equation}
a_s = softmax( A_g ),
\end{equation}
which forces either $a_0$ or $a_1$ to a $0$ value once the training converges \cite{Liu2018b}.

\subsubsection{Loss Definition}
As introduced before, additional objectives/constraints related to the final HW characteristics may lead to non convergence
issues (see Section~\ref{sec:unipolar_weights}).
In order to help the convergence towards a valid solution, we introduce
extra $\mathcal{L}_C$ terms in the loss computation that may depend on the
training step.
The final loss $\mathcal{L}_F$ is then defined as
\begin{equation}
\mathcal{L}_F = \mathcal{L} + \mathcal{L}_{L2} + \mathcal{L}_{L1} + \mathcal{L}_{C},
\end{equation}
where $\mathcal{L}$ refers the standard training loss,
$\{\mathcal{L}_{L1}, \mathcal{L}_{L2}\}$ refer the standard $L1$ and $L2$ regularization losses,
and $\mathcal{L}_C$ is the custom penalization.
An example of this particular regularization terms may refer the penalization
of weight values beyond a threshold $W_T$ after training step $N$.
This loss term can be formulated as
\begin{equation}
\mathcal{L}_C = \alpha_C \sum_{w} max(W-W_T, 0) HV(step - N)
\label{eq:loss_c}
\end{equation}
where $\alpha_C$ is a preset constant and $HV$ the \emph{Heaviside} function.
If the training would still provide weights whose values surpass $W_T$,
$HV$ function can be substituted by a non clipped function $relu(step-N)$.
In particular, this $\mathcal{L}_C$ function was used in the unipolarity experiments located
at Section~\ref{sec:results}.

\subsubsection{Implemented Quantization Scheme}

The implemented quantization stage takes as input a random tensor $T = \{t_t\}, t_t \in \mathbb{R}$
and projects it to the quantized space $Q = \{q_{q+}, q_{q-}\}$,
where $q_{q+} = \alpha_Q 2^{q}$, $q_{q-} = -\alpha_Q 2^{q}$, and $\alpha \in \mathbb{R}$.
Therefore the projection is denoted as $q(T) = T_q$, where $T_q = \{t_q\}, t_q \in Q$.
For its implementation we use \emph{fake\_quant} operations \cite{Krishnamoorthi2018}
computing \emph{straight through estimator} as the quantization scheme,
which provides us with the uniformly distributed $Q$ set, always including $0$.
However, the quantization nodes shown in Figure~\ref{fig:training_quant_graph}
allow the use of non-uniform quantization schemes.
The definition of the quantized space $Q$ gets determined by the minimum and
maximum values given by the global variables $V_g$.

Algorithm~\ref{alg:darts} can consider either $max/min$ functions or
stochastic quantization schemes \cite{Krishnamoorthi2018}. 
Similarly, the quantization stage is dynamically activated/deactivated using the
global variable $do_Q \in {0, 1}$,
with could be easily substituted to support incremental approaches \cite{Zhou2017a}.
In particular, and as shown in Section~\ref{sec:unipolar_weights},
the use of \emph{alpha-blending} scheme \cite{Liu2019} proves useful when the weight precision is very limited.

\subsection{Unipolar Weight Matrices Quantized Training}
\label{sec:unipolar_weights}
Mapping positive/negative weights
to the same crossbar involve double the crossbar resources and introducing additional periphery.
Using the proposed training scheme we can restrict further the characteristics
of the DNN graph obtaining unipolar weight matrices, by redefining some global variables as
\begin{equation}
W_g  \in [0, w_1]
\end{equation}
and introducing the $\mathcal{L}_C$ function defined by Equation~\ref{eq:loss_c}.

Moreover, for certain activations (\emph{relu}, \emph{tanh}, etc.)
the maximum and/or minimum values are already known, and so the sets
of parameters in $V_g$ can be constrained even further.
These maximum and minimum values can easily be mapped to specific parameters
in the activation function circuit interfacing the crossbar \cite{Giordano2019}.
Finally, in cases where weights precision is very limited (i.e. $2$ bits),
additional loss terms as $\mathcal{L}_C$ gradually move weight distributions
from a bipolar space to an only positive space, helping the training to converge.

In summary, by applying the mechanisms described in Section~\ref{sec:hw_aware},
we open the possibility of obtaining NN graphs only containing unipolar weights.

\section{Experiments and Results}
\label{sec:results}

\begin{figure}[!t]
\centering
\includegraphics[width=0.75\columnwidth]{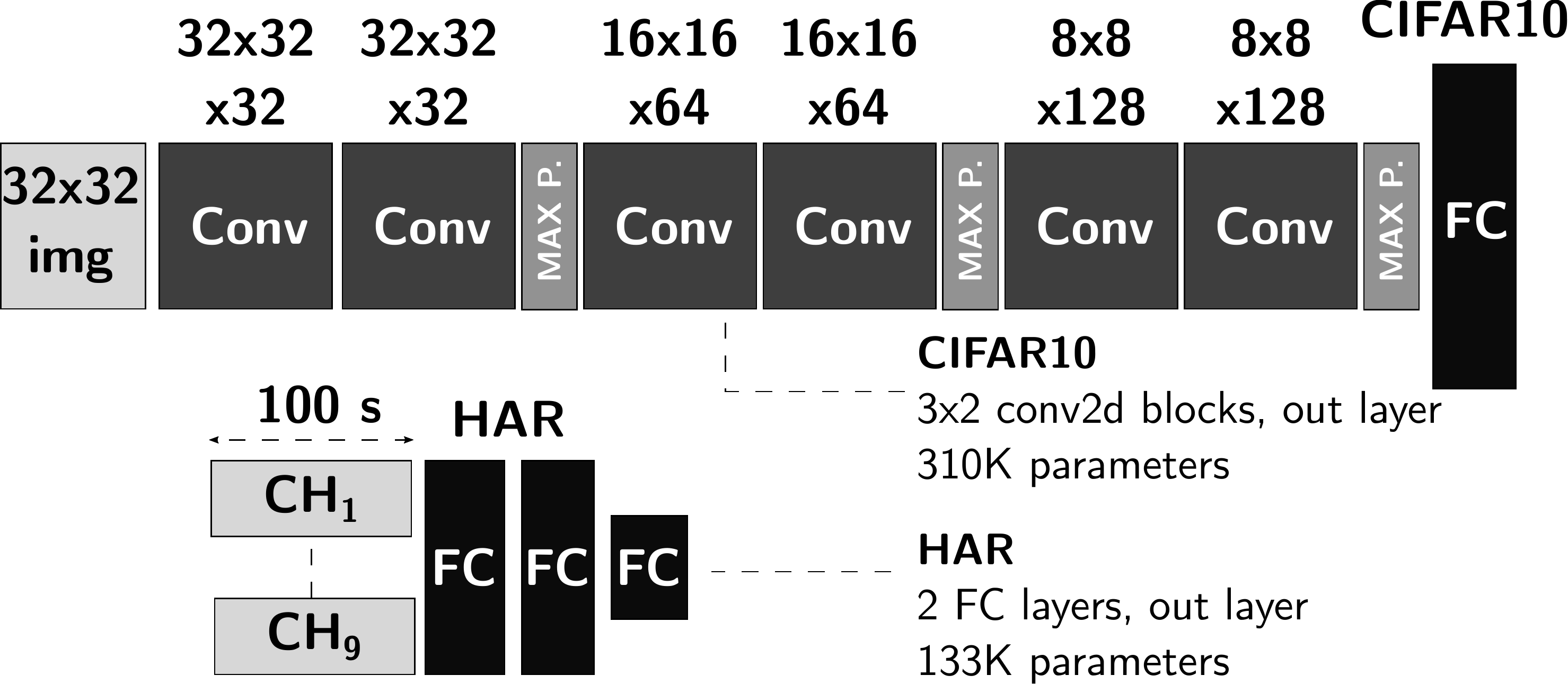}
\caption{
	Structure of the CIFAR10 and HAR classification NNs.
}
\label{fig:nn}
\end{figure}

We have evaluated the presented methodology using
CIFAR10 and Human Activity Recognition (HAR) applications.
CIFAR10 \cite{Krizhevsky2009} comprises the classification of
$32x32$ sized images into $10$ different categories.
HAR classifies among incoming data from different sensors (accelerometer,
gyroscope, magnetometer, $3$ channels each) into $12$ different activities (run, jump, etc.)
To mimic a smartwatch scenario we used real data from sensors placed in only one limb from
\cite{Banos2014} dataset.

Figure~\ref{fig:nn} describes the architectures used in each case:
CIFAR10 problem represents a good example of always-ON medium sized DNNs,
including multiple convolutional layers and $310K$ parameters.
HAR NN interfaces $9$ input channels, with a time-series data input of $100$ samples each,
followed by 2 fully connected layers with a total of $133K$ parameters.

\subsection{Accuracy vs Uniform Scaling Trade-off Results: CIFAR10}

After performing a quantization hyperparameter design exploration
we conducted the quantized training of the use case NN using both the standard
STE approach in \emph{TensorFlow} library \cite{Krishnamoorthi2018} and the proposed scheme.
It is to be noted that \emph{TensorFlow's} scheme does not quantize the bias.
and thus when mapped to the crossbars, additional quantization studies would be needed.
\begin{figure}[!t]
\centering
\includegraphics[width=\columnwidth]{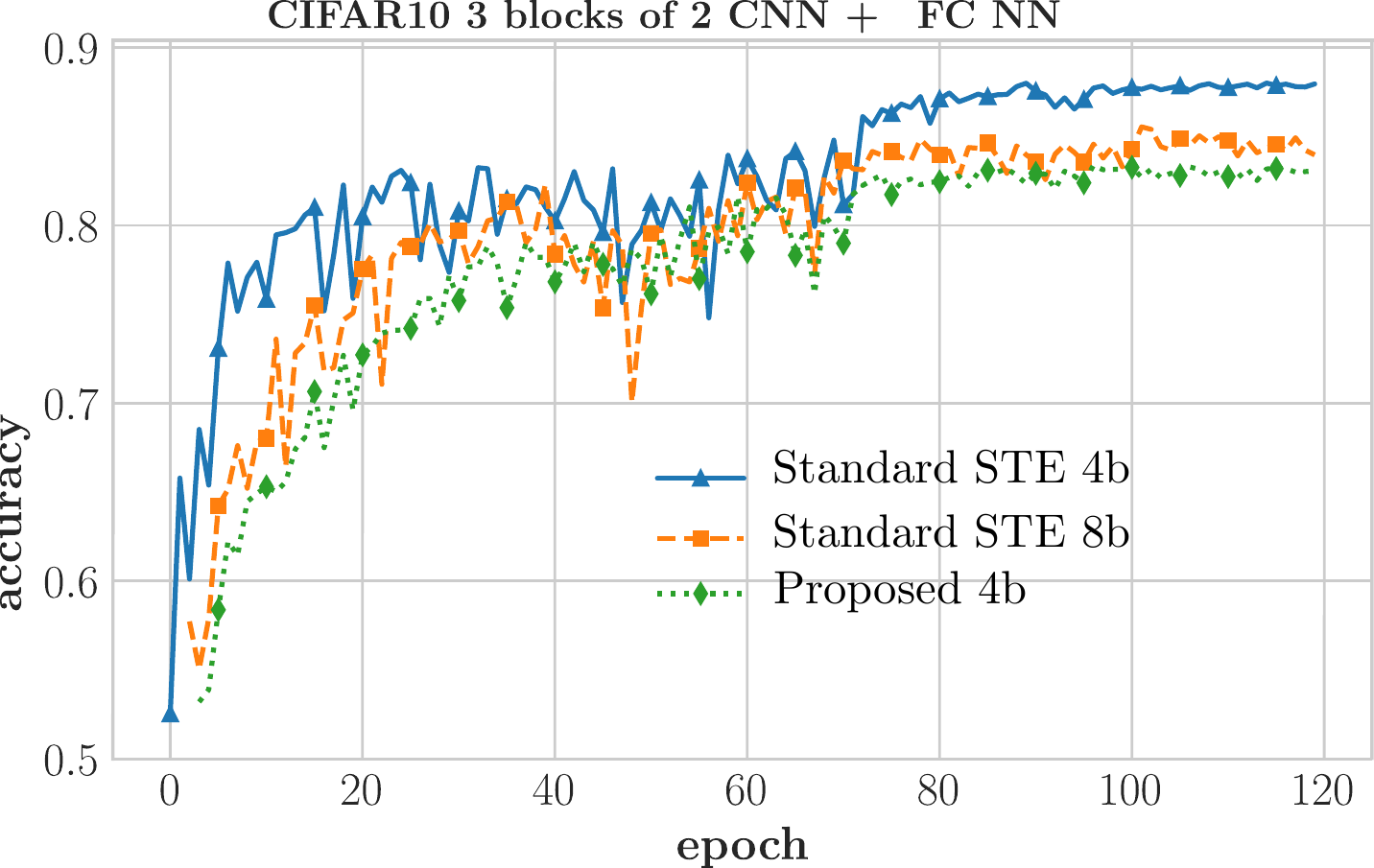}
\caption{
	CIFAR10 CNN. Comparison between $4$-bit quantized training
	with \emph{TensorFlow} STE quantization and the proposed solution.
	$8$-bit is shown as a baseline.
}
\label{fig:results_cifar_1}
\end{figure}
\begin{table}[t]
\caption {CIFAR10 CNN quantization schemes comparison.
Our proposal brings a $5.7\times$ reduction on
the number of different weights.
}
\label{tab:cifar}
\vskip 0.15in
\begin{center}
\begin{footnotesize}
\begin{tabular}{lccc}
	\toprule


	\textbf{Scheme} & \textbf{Accuracy} & \textbf{\# Different} & \textbf{Uniform} \\
	\textbf{} & \textbf{} & \textbf{Weights} & \textbf{DACs/ADCs} \\
	\midrule
	TF, $8$-bit & $88.10\%$ & $1372$ & No \\
	TF, $4$-bit & $84.43\%$	& $91$   & No \\
	Proposed, $4$-bit   & $83.7\%$  & $16$   & Yes \\
	\bottomrule
\end{tabular}

\end{footnotesize}
\end{center}
\vskip -0.1in
\end{table}
Figure~\ref{fig:results_cifar_1} shows the evolution of the Deep convolutional NN
learning through the training process.
When quantized with $4$-bit (weights and activations) our solution
gives accuracies only $0.7\%$ away of the state of the art.
Moreover, and as described in Table~\ref{tab:cifar} our solution provides
a significant reduction in the number of full-custom circuit modules involved in the
algorithm-to-HW mapping.

\subsection{Unipolar Weights vs Accuracy Trade-off}
\subsubsection{FC DNN: HAR}
In this experiment we apply the proposed mechanisms to obtain a NN
classifying different HAR activities whose weights take only positive values.
The DA Algorithm~\ref{alg:darts} conducted the exploration of the NN design space,
determining the NN architecture and parameters set that
provided the best accuracy while using only positive weights within the NN.
To help the NN training to converge, and following graph structure
shown in Figure~\ref{fig:training_quant_graph}, custom regularizers were required
to penalize negative weights, and a variation of \emph{alpha-blending} quantization scheme \cite{Liu2019} was introduced.
\begin{figure}[!t]
\centering
\includegraphics[width=\columnwidth]{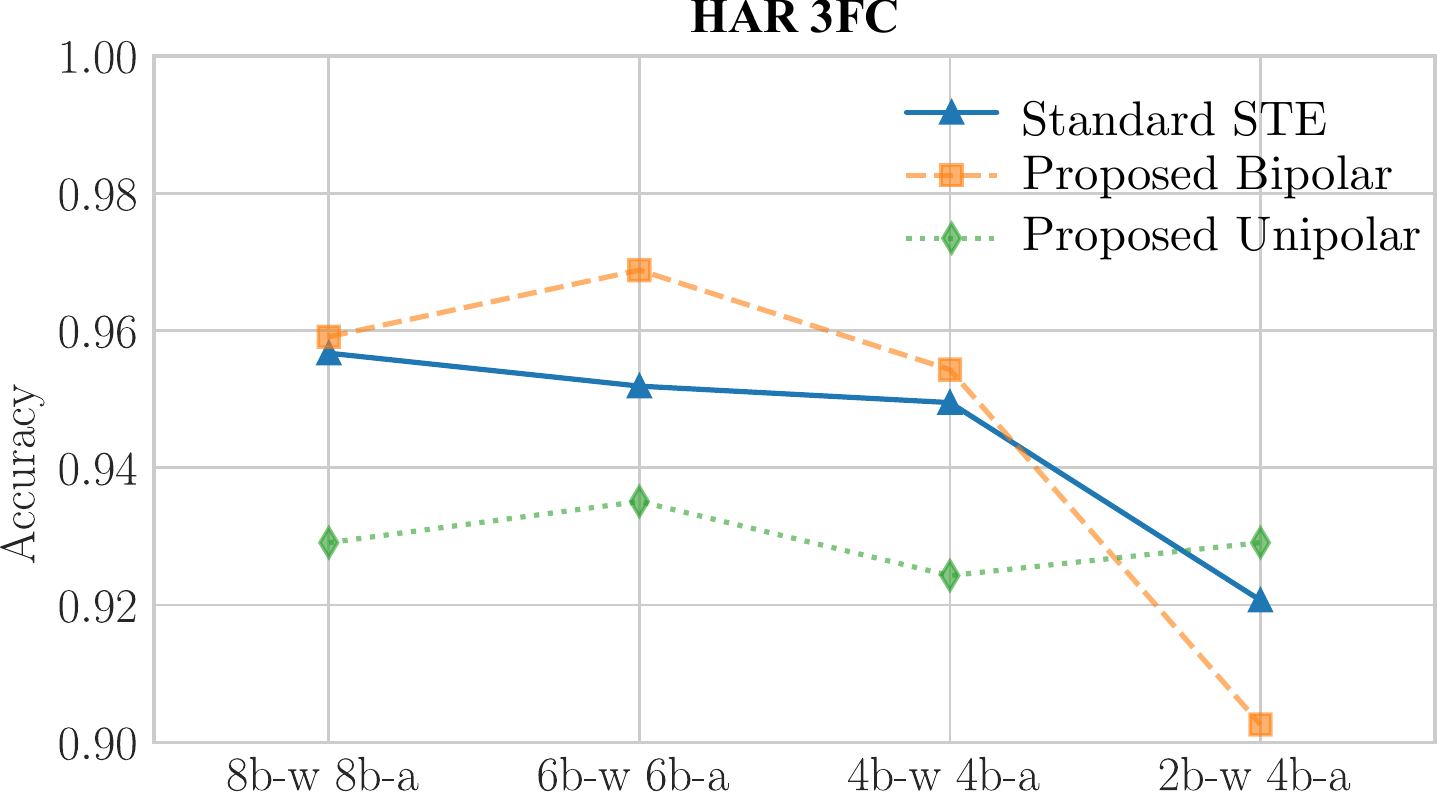}
\caption{
	Comparison between \emph{TensorFlow} STE quantization
	and proposed solution. Accuracy levels are indistinguishable
	even with unipolar weight matrices.
}
\label{fig:results_har_1}
\end{figure}
Figure~\ref{fig:results_har_1} summarizes the experiment results.
NNs with bipolar weight matrices use \emph{relu} as the hidden-layers activation.
On the contrary, our design exploration algorithm found $act = tanh(x - th_g)$ as
a function best suiting NNs based on unipolar weight matrices.
The introduction of $th_g$ shift on the \emph{tanh} function allows the network to
map a small valued positive (negative) input to the activation as a small valued
negative (positive) at the output, .
The proposed solution is as competitive as the standard one,
while obtaining the significant benefit of a reduced set of weights and uniform HW.
But more importantly, we demonstrate that small NN using only unipolar weight matrices
/ADC
can correctly perform classifications, aiding the deployment to NVM crossbars.

\subsubsection{Deeper CNN: CIFAR10}

\begin{figure}[!t]

\centering
\includegraphics[width=\columnwidth]{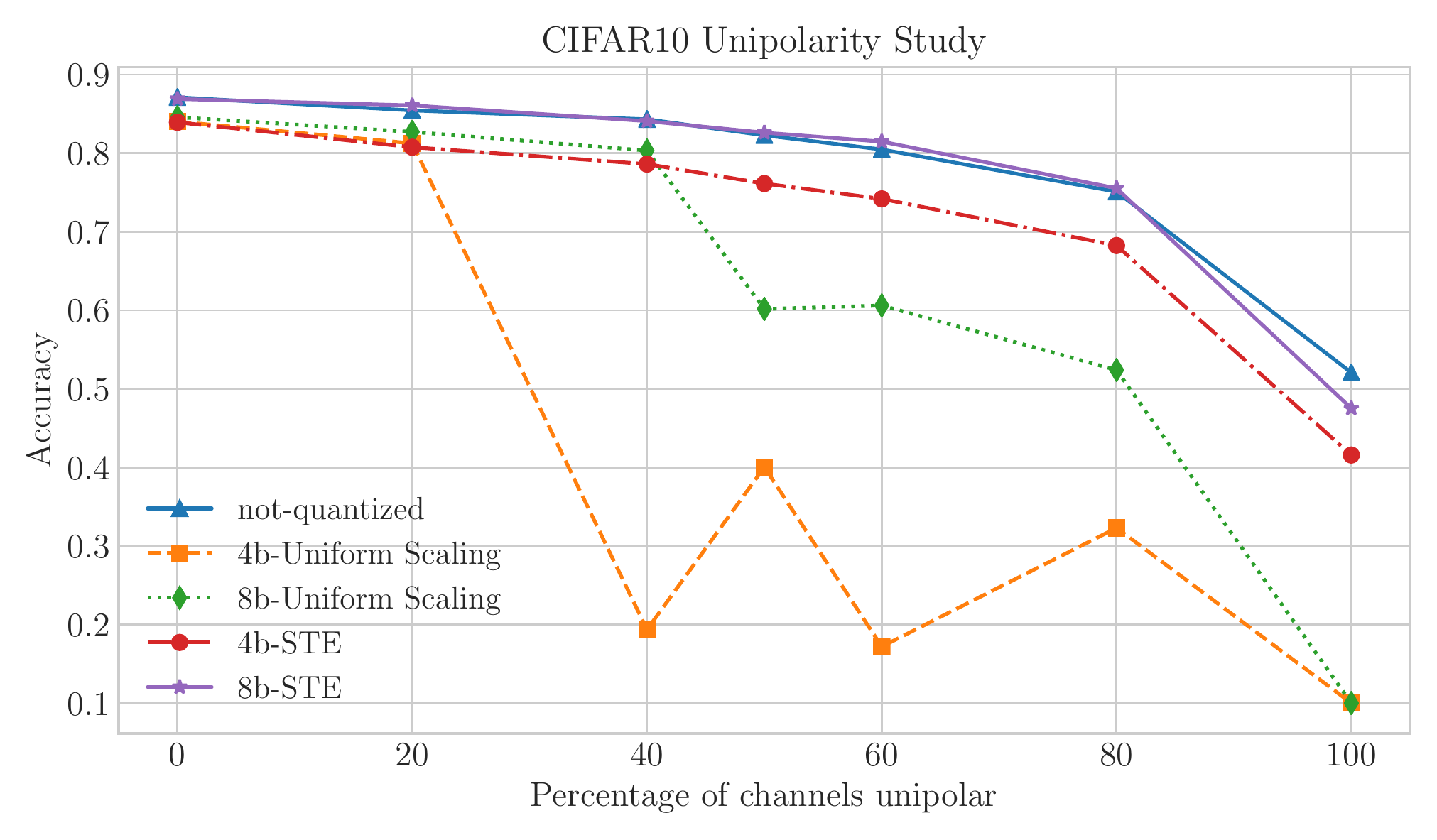}
\caption{
	CIFAR10 NN study on variable percent of unipolar channels.
	Reasonable accuracies are achieved with $40-50\%$ of bipolar weights.
}
\label{fig:unipolar_cifar10}
\end{figure}
For larger convolutional networks the imposed unipolarity constraint can be
to restrictive for the NN to correctly learn. 
We propose imposing the constraint only to a certain amount of channels in each layer.
The ratio of unipolar/bipolar channels will determine the final accuracy and the 
power and area savings.

Figure~\ref{fig:unipolar_cifar10} describes the results of applying hard unipolar weights constraint
to the same CIFAR10 application, varying the percentage of unipolar channels in each 
convolutional stage from $0\%$ (bipolar weights) to $100\%$ (completely unipolar weights), for the standard STE and uniform-scaling quantization approaches.
It can be seen how a minimum number of channels in each convolutional layer
is required by the NN to learn.
Unipolar percentages above $60\%$ impose a hard limitation, specially when the uniform-scaling training scheme is used.
However, it can be seen how for the $8$-b STE quantization scheme, by imposing $50\%$ unipolar channels, we can reduce a $25\%$ the crossbar area/energy with a small $2\%$ accuracy penalty.
For the $4$-b scheme, a $20\%$ area/energy savings would come with a $4\%$ accuracy reduction.
Therefore we can state that even for more complex problems, 
we can greatly simplify the NN deployment forcing a percentage of channels to be unipolar.

\subsection{Energy and Area Benefits}
\begin{figure}[!t]
\centering
\includegraphics[width=\columnwidth]{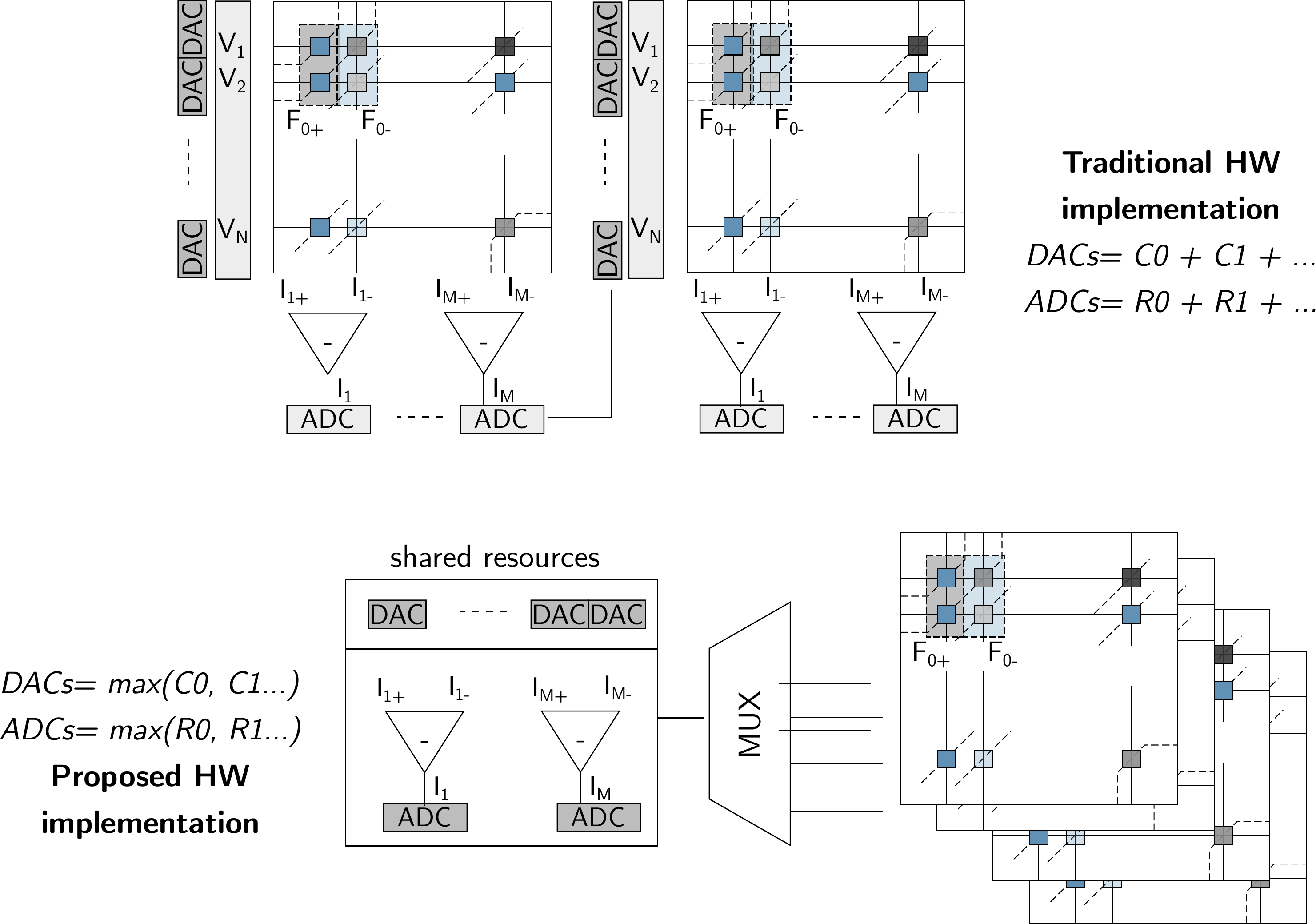}
\caption{
	HW implementation differences between traditional and proposed approach,
	highlighting the saved periphery by using a multiplexing scheme.
}
\label{fig:proposal_hw}
\end{figure}
Figure~\ref{fig:proposal_hw} describes the comparison of HW implementation,
where we consider \cite{Hamdioui2019},
in which each PCM NVM element --each parameter in our NN-- consumes $~0.2$ $\mu$W,
and has a $25F^2$ area, equivalent to $0.075 \mu m^2$.
For the DAC/ADC characteristics, we designed in house $4$-bit and $8$-bit
elements, using a $55$ nm CMOS technology.
Simulated power consumption and area are gathered in Table~\ref{tab:dac_adc}.
\begin{table}[!t]
\centering
\begin{footnotesize}
\caption {Characteristics of designed ADC and DAC}
\begin{tabular}{l c c c}
	\toprule
	\textbf{Device} & \textbf{Power@$10$ MHz} & \textbf{@$100$ MHz} & \textbf{Area} \\
	\midrule
	DAC $4$b & $3.2$ $\mu$W & $11.7$ $\mu$W & $101$ $\mu$m$^2$ \\
	DAC $8$b & $4.4$ $\mu$W & $13.6$ $\mu$W & $440$ $\mu$m$^2$ \\
	ADC $4$b & $1.28$ $\mu$W & $12.56$ $\mu$W & $1030$ $\mu$m$^2$ \\
	ADC $8$b & $1.64$ $\mu$W & $16.39$ $\mu$W & $7920$ $\mu$m$^2$ \\
	\bottomrule
\end{tabular}
\label{tab:dac_adc}
\end{footnotesize}
\end{table}
A power/area overhead of $5/10 \%$ over the ADC figure for an integrated adapted current subtractor is added in the case where bipolar weights are present. Additional $5 \%$ power penalty is applied for ADCs using current scaling.
Regarding each one of the NN layers, DACs and ADCs will only be multiplexed should the layer
maintain uniform scaling with the system.
With our proposed approach, all layers share the same input ranges, and only
the last layer ADCs would be different from the rest of the system.

\subsubsection{Energy Estimation}

To maximize the throughput per NN layer
we consider one DAC (ADC) per column (row).
From the power perspective, for each layer the total number of NVM cell reads performing the multiplications
(and automatically the additions) would be $\sum_{F_i} \mathcal{K}_i \mathcal{X}_i$
for the convolutional layers and $\mathcal{X}_i \mathcal{Y}_i$ for the fully connected ones,
where $\mathcal{X}_i, \mathcal{Y}_i, \mathcal{K}_i$ refer the the size of inputs, outputs,
and kernel respectively, and $F_i$ refers the number of filters of the $i-th$ layer.
Regarding the DACs and ADCs utilization, a total of $\mathcal{X}_i$ and digital to analog
and $\mathcal{Y}_i F_i$ analog to digital conversions are required.
No analog scaling system is required.
The results describing the power estimation per inference in both bipolar-CIFAR10 and
bipolar/unipolar-HAR applications is displayed in Table~\ref{tab:power_estimation}.
It can be seen that as bipolar weights were needed in the image solution,
and due to the amount of multiplications ($>38$ million per inference),
the saved power is almost negligible.
However, in very low power IoT applications, the proposed solution requires only $55\%$ of
the energy compared with traditional schemes, mainly due to the unipolar
weight matrices encoded in the NVM crossbar.
\begin{table}[!t]
\centering
\caption {Estimated energy per inference:
	number of NVM cell reads (positive $(+)$ and negative$(-)$ weights),
	DAC/ADC operations.
}
\begin{footnotesize}
\begin{tabular}{l c c c}
	\toprule
	\textbf{CIFAR10} & \textbf{TF $8$ bits} & \textbf{TF $4$ bits} & \textbf{Ours $4$ bits} \\
	\textbf{} & \textbf{10/100 MHz} & \textbf{10/100 MHz} & \textbf{10/100 MHz} \\
	\midrule
	\textbf{Total} & $1.6 / 0.19$ $\mu$J  & $1.59 / 0.18$ $\mu$J  & $1.58 / 0.18$ $\mu$J \\
	NVM ($\pm$) $\approx 77e6$ & $1.55 / 0.16$ $\mu$J & $1.55 / 0.16$ $\mu$J & $1.55 / 0.16$ $\mu$J \\
	DAC ops $\approx  75e3$   & $32.9 / 10.1$ nJ & $23.9 /  8.7$ nJ & $23.9 /  8.7$ nJ \\
	ADC* ops $\approx  115e3$ & $22.6 / 22.6$ nJ & $18.4 / 18.1$ nJ & $16.5 / 16.2$ nJ \\
	\midrule
	\textbf{HAR} & \textbf{TF $8$ bits} & \textbf{TF $4$ bits} & \textbf{Ours $4$ bits} \\
	\textbf{} & \textbf{10/100 MHz} & \textbf{10/100 MHz} & \textbf{10/100 MHz} \\
	\midrule
	\textbf{Total}  & $1.6 / 0.24$ nJ  & $1.54 / 0.22$ nJ  & $0.84 / 0.15$ nJ \\
	NVM ($+$) $\approx 34e3$ & $0.7 / 0.07$ nJ & $0.7 / 0.07$ nJ & $0.7 / 0.07$ nJ \\
	NVM ($-$) $\approx 34e3$ & $0.7 / 0.07$ nJ & $0.7 / 0.07$ nJ & $0$ nJ \\
	DAC ops $\approx  384$   & $0.17  / 0.05$ nJ & $0.12 / 0.04$ nJ & $0.12 / 0.04$ nJ \\
	ADC* ops $\approx  268$  & $0.052 / 0.05$ nJ & $0.04 / 0.03$ nJ & $0.03 / 0.03$ nJ \\
	\bottomrule
\end{tabular}
\end{footnotesize}
\label{tab:power_estimation}
\end{table}

\subsubsection{Area Estimation}
In traditional deployments, being $F_i$ the number of filters present in 
a given layer $L_i$,
$F_i$ full custom different ADCs would be designed and placed for that layer, 
freezing the applicability to a particular application.
However, with our proposed scheme we can deploy different NN applications
in the same hardware, using many smaller and fixed-sized crossbars.
We can feed the incoming inputs in batches, reusing the kernels unrolled in the crossbar.
Adopting this second scheme for the CIFAR10 example, the largest CNN unrolled layer requires
an input of size $32x32x32$. For example, if the crossbar size available in our
reconfigurable system is $128x128$, the layer can be batched in $256$ operations.
If the hardware blocks were composed of $512x128$ elements, the layer could be batched in $64$ operations.
On the other hand, for smaller NN this same hardware could fit entire layers: in HAR benchmarch each layer can fit in a $128x128$ crossbar.
For both crossbar size examples, every layer but the last would reuse the $128$ DAC/ADC pairs during inference.

Table~\ref{tab:area_estimation} summarizes the area estimation 
when considering crossbars composed of $128x128$ elements
(a very conservative approach to avoid technology problems) and
assisted by $128$ DACs, $128$ ADCs and additional periphery.
For the traditional approaches, we follow the deployment schemes in the literature,
and consider that the number of ADCs present in each layer
does not need to match the crossbar column size, saving considerable amount of area
but avoiding reconfigurability.
On the contrary, by using the proposed solution the DACs and ADCs are multiplexed.
\begin{table}[!t]
\centering
\begin{footnotesize}
\caption {Estimated area using
	$128x128$ basic crossbar blocks.}
\begin{tabular}{l c c c}
	\toprule
	\textbf{CIFAR10} & \textbf{TF $8$ bits} & \textbf{TF $4$ bits} & \textbf{Ours $4$ bits} \\
	\midrule
	Reconfigurable & No & No & Yes \\
	\midrule
	Crossbars & $44$ & $44$ & $44$ \\
	DACs & $448$ & $448$ & $128$ \\
	ADCs & $896$ & $896$ & $256$ \\
	Current subtractors & $896$ & $896$ & $256$ \\
	\midrule
	Total Area & $8.05$ mm$^2$  & $1.1$ mm$^2$  & $0.22$ mm$^2$ \\
	\midrule
	\textbf{HAR} & \textbf{TF $8$ bits} & \textbf{TF $4$ bits} & \textbf{Ours $4$ bits} \\
	\midrule
	Reconfigurable & No & No & Yes \\
	\midrule
	Crossbars & $6$ & $6$ & $3$ \\
	DACs & $384$ & $384$ & $128$ \\
	ADCs & $268$ & $268$ & $256$ \\
	Current subtractors & $268$& $268$& $0$\\
	\midrule
	Total Area & $2.51$ mm$^2$  & $0.35$ mm$^2$ & $0.28$ mm$^2$ \\
	\bottomrule
\end{tabular}
\label{tab:area_estimation}
\end{footnotesize}
\end{table}
The benefits are noticeable:
First, we guarantee that the HW is uniform across the NN, ensuring reconfigurability.
Second, in CIFAR10 benchmark, the solution leads to up to $80\%$ area saving
--$0.22$ mm$^2$ vs $1.1$ mm$^2$ for $4$ bit accelerators.
For HAR benchmark, up to $20\%$ area saving is achieved.
When comparing against the traditional $8$ bit deployment schemes, this area savings raise up to $97\%$ for CIFAR10 CNN benchmark and $89\%$ for the HAR NN.

\section{Conclusions}
\label{sec:conclusions}

ML at the edge requires accelerators that efficiently compute inference
in constrained devices, and NVM based analog accelerators are promising candidates
due to their low power capabilities.
However the full-custom per-layer design of the periphery interacting with the crossbars
hinder the reconfigurability of the whole system.

This work has presented the first solution that aids
the algorithm deployment in uniform crossbar/periphery blocks, at training time.
With no accuracy penalty, the method is able to simplify
the design of the crossbar periphery, significantly
reducing the overall area and power consumption,
and enabling real re-usability and reconfigurability.
Moreover, we have demonstrated that DNN with unipolar weight matrices
can correctly perform bio-signals classification tasks
while solving the negative/positive weights problem inherent to NVM crossbars,
and therefore reducing by half the crossbar area/energy and significantly simplifying
the periphery design. We validated our solution against two different always-ON
sensing applications, CIFAR10 and HAR, obtaining competitive accuracies
while simplifying the whole system design.

\section{Acknowledgements}
This research on CIM architecture is supported by EC Horizon 2020
Research and Innovation Program through MNEMOSENE project under Grant
780215.


%

\bibliography{ijcnn_2020}
\bibliographystyle{noUrlIEEEtran}

\end{document}